\documentclass[letterpaper, paper,11pt]{AAS}		

\usepackage{bm}
\usepackage{amsmath, amssymb}
\usepackage{subfigure}
\usepackage[colorlinks=true, pdfstartview=FitV, linkcolor=black, citecolor= black, urlcolor= black]{hyperref}
\usepackage{overcite}
\usepackage{footnpag}			      	
\usepackage{soul, color}
\usepackage{booktabs}    

\usepackage[symbol]{footmisc}

\DeclareMathOperator*{\argmin}{arg\,min}
\DeclareMathOperator*{\argmax}{arg\,max}

\PaperNumber{21-654}

\begin{document}

\title{Robotic Testbed for Rendezvous and Optical Navigation: Multi-Source Calibration and Machine Learning Use Cases}

\author{Tae Ha Park\thanks{Ph.~D.~Candidate, Department of Aeronautics \& Astronautics, Stanford University, Stanford, CA 94305.}, 
Juergen Bosse\thanks{Managing Director, Robo-Technology GmbH, Benzstrasse 12, D-82178 Puchheim, Germany.},
Simone D'Amico\thanks{Associate Professor,  Department of Aeronautics \& Astronautics, Stanford University, Stanford, CA 94305.}
}

\maketitle{} 		

\begin{abstract}
This work presents the most recent advances of the Robotic Testbed for Rendezvous and Optical Navigation (TRON) at Stanford University - the first robotic testbed capable of validating machine learning algorithms for spaceborne optical navigation. The TRON facility consists of two 6 degrees-of-freedom KUKA robot arms and a set of Vicon motion track cameras to reconfigure an arbitrary relative pose between a camera and a target mockup model. The facility includes multiple Earth albedo light boxes and a sun lamp to recreate the high-fidelity spaceborne illumination conditions. After the overview of the facility, this work details the multi-source calibration procedure which enables the estimation of the relative pose between the object and the camera with millimeter-level position and millidegree-level orientation accuracies. Finally, a comparative analysis of the synthetic and TRON simulated imageries is performed using a Convolutional Neural Network (CNN) pre-trained on the synthetic images. The result shows a considerable gap in the CNN's performance, suggesting the TRON simulated images can be used to validate the robustness of any machine learning algorithms trained on more easily accessible synthetic imagery from computer graphics.
\end{abstract}

\section{Introduction}
The vision-only navigation of a spacecraft about noncooperative Resident Space Objects (RSO) is an enabling technology for future on-orbit servicing and debris removal missions. Unlike those based on complex sensors such as Light Detection and Ranging (LIDAR) or stereovision, monocular navigation systems utilize a commercially available low Size, Weight, Power, and Cost (SWaP-C) camera, making it an attractive choice of sensor due to its low mass and power requirements. The key component of monocular navigation is to determine the pose (i.e., position and orientation) of the target relative to the servicer's camera based on a single or a sequence of images. The conventional approach is to first extract and process salient features such as points\cite{Harris1988Corner}, edges\cite{Canny1986, Ballard1981HoughTransform}, scale-invariant features such as SIFT\cite{Lowe2004SIFT}, SURF\cite{Bay2008SURF} and ORB\cite{Rublee2011ORB} features of a spacecraft, or landmark features such as craters of an asteroid\cite{Dennison2021SFM}. These features are then compared with those of the available target 3D model to compute the 6D pose\cite{Damico2014IJSSE, Sharma2018JSR, Capuano2019RobustFeatures, Capuano2018Pose, Sharma2017AASGCC, Dennison2021SFM}. Recently, Machine Learning (ML) techniques based on Convolutional Neural Networks (CNN) have been developed to replace the feature extraction step with superior performance\cite{Sharma2019AAS, Park2019AAS, Kisantal2020SPEC, PasqualettoCassinis2020CNNEKF, Chen2019SatellitePE}. However, unlike image processing-based methods, a CNN must be trained on a large set of target images with accurate pose labels. While the data-hungry nature of the training is a ubiquitous challenge for any ML applications, it is especially difficult and outright impractical in spaceborne applications to acquire a large quantity of images of interested targets in various space environments with accurate pose labels. Therefore, existing works rely on synthetic images generated from a computer graphics renderer such as OpenGL\cite{Sharma2019AAS, Kisantal2020SPEC, Park2019AAS, Chen2019SatellitePE} or Blender\cite{Capuano2018Pose} to train and test the CNN subsystems.

Unfortunately, the synthetic renderers cannot faithfully replicate various illumination and noise artifacts present in the spaceborne imagery. Naturally, CNNs trained with synthetic images alone would overfit to the features inherent to the synthetic imagery and thus have degraded performance on the spaceborne imagery\cite{Sharma2019PhDThesis, Kisantal2020SPEC, Park2019AAS, Chen2019SatellitePE}. Therefore, in order for a CNN-based system to be deployed to space missions: 1) it must be trained to be robust against various adversarial conditions in space such as high contrast, extreme shadowing and low signal-to-noise ratio, and 2) its robustness must be validated on ground with no access to target spaceborne images. Previous works have attempted to address the robust training with extensive data augmentation during training\cite{Park2019AAS, PasqualettoCassinis2020CNNEKF}, but relatively fewer works or efforts have been dedicated to the issue of on-ground validation of CNNs\cite{Sharma2019AAS}. The most promising method involves a robotic testbed that is capable of re-creating various space environmental conditions and configuring the camera and the target model to achieve the desired relative pose with high accuracy. Such facility would allow one to obtain the quasi-spaceborne imagery of an arbitrary quantity and characteristics with statistical distribution completely different from synthetic images. Such an imagery can then be used to evaluate the robustness of the CNN trained with synthetic images or any other vision-based navigation algorithms developed based on them.

\begin{figure}[!t]
	\includegraphics[width=0.6\textwidth]{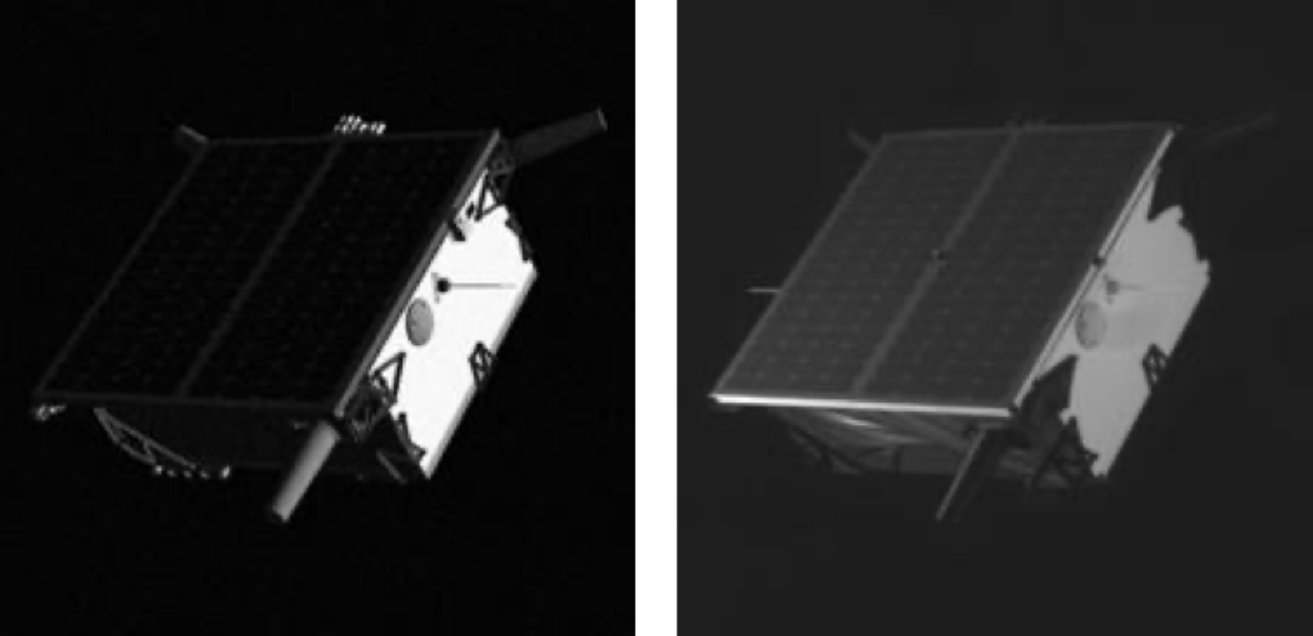}
	\centering
	\caption{SPEED synthetic (\emph{left}) and simulated (\emph{right}) cropped images with the pose label estimated from the TRON testbed.}
	\label{fig:SPEED}
\end{figure}

\begin{figure}[!t]
	\includegraphics[width=0.7\textwidth]{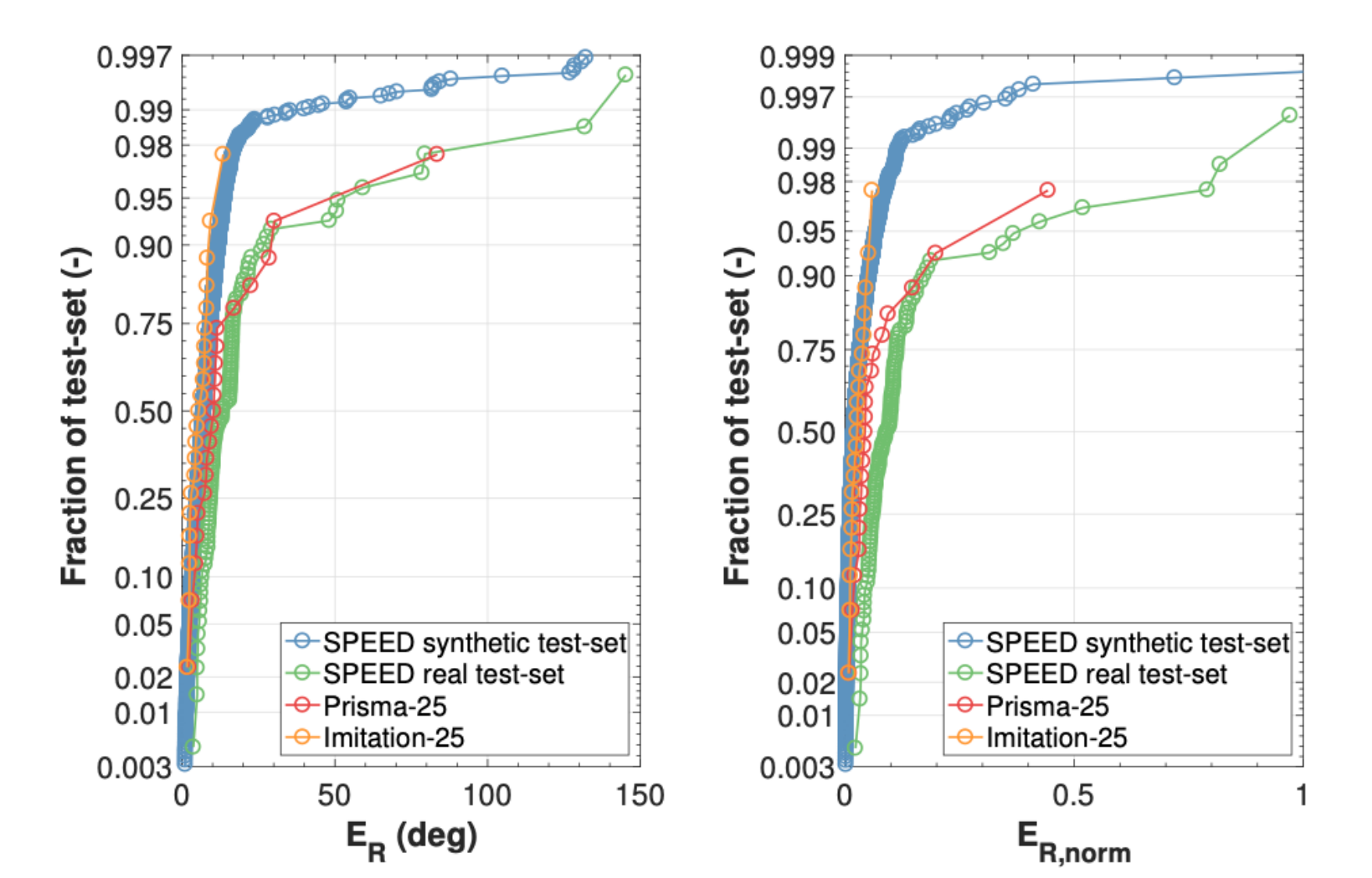}
	\centering
	\caption{(\emph{Left}) Cumulative distribution of rotation error of SPN\cite{Sharma2019AAS} when tested on SPEED synthetic test set, real test set, PRISMA-25, and Imitiation-25 datasets. (\emph{Right}) Same distribution of the rotation error normalized by inter-spacecraft separation and camera field-of-view. Figures taken from Ref [\citen{Sharma2019PhDThesis}].}
	\label{fig:sharma spn error}
\end{figure}

One example of such approach is the Spacecraft Pose Estimation Dataset (SPEED)\cite{Sharma2019SPEEDonSDR}, which was made publicly available in 2019. SPEED contains 15,000 synthetic images of the Tango spacecraft from the PRISMA mission\cite{PRISMA_chapter, Damico2014IJSSE} and 300 simulated images of the full-scale Tango mockup model captured from the Robotic Testbed for Rendezvous and Optical Navigation (TRON) at the Stanford's Space Rendezvous Laboratory (SLAB). As seen in Figure \ref{fig:SPEED}, a simulated image describes identical geometric features of the target yet has fundamentally different visual characteristics compared to its synthetic counterpart. In fact, the result of the Satellite Pose Estimation Competition (SPEC), co-hosted by SLAB and the Advanced Concepts Team (ACT) of the European Space Agency (ESA) in 2019, shows virtually all the top-performing CNNs trained on synthetic images have degraded performances on simulated images\cite{Kisantal2020SPEC}. The similar trend is also described in Figure \ref{fig:sharma spn error}, which shows that the Spacecraft Pose Network (SPN) model trained on SPEED synthetic images have worse attitude predictions on SPEED simulated (real) test set and PRISMA-25, which consists of 25 spaceborne images from the rendezvous phase of the PRISMA mission. Interestingly, the degraded yet comparable performances on the simulated and spaceborne images suggest that the simulated images from a robotic testbed with the capabilities of TRON can thus be used to evaluate the robustness of a CNN for spaceborne applications. However, the simulated imagery of SPEED is currently restricted to 300 images with extremely restricted pose distribution and variety of illumination conditions. Therefore, a significant upgrade must be made to validate the performance on a wide range of navigation scenarios.

Several other laboratories have constructed similar testbeds to simulate vision-only closed-loop navigation and control algorithms. Some examples include ASTROS at the Georgia Institute of Technology\cite{Tsiotras2014ASTROS, Dor2018ORBSLAM}, POSEIDYN at the Naval Postgraduate School\cite{Zappulla2017POSEIDYN}, and M-STAR at California Institute of Technology\cite{Nakka2018MSTAR}. These facilities commonly employ air-bearing platforms on a flat epoxy or granite floor with thrusters and actuators to simulate the spacecraft movement and Vicon motion-tracking cameras to provide ground-truth pose labels. While these testbeds excel at simulating actual spacecraft movement given maneuver commands, none of them are tailored for ML applications, as the capability of efficiently reconfiguring the arbitrary pose commands at large quantity has never been showcased. Recently, the GNC Rendezvous, Approach and Landing Simulator (GRALS) testbed at the European Space Research and Technology Centre (ESTEC), a facility comprising a ceiling-mounted KUKA robotic arm and Vicon motion track cameras, was used to generate 100 simulated images of 1:25 mockup of the Envisat satellite\cite{Cassinis2021ORGL}. However, similar to the earlier generation of TRON, the target is mounted on a static tripod, severely restricting the image acquisition from different viewpoints. 

The first contribution of this paper is the introduction of the next generation TRON facility at SLAB, the first testbed capable of accurately reconfiguring an arbitrary relative pose with high-fidelity space-like illumination conditions. Unlike other facilities, TRON includes two KUKA 6 degrees-of-freedom (DOF) robot arms\cite{KUKA} respectively holding a camera and a lightweight, reduced-scale model of the target RSO (see Figure \ref{fig:TRON Simulation Room}). One robot is installed on a ceiling-mounted linear rail running through the facility; therefore, compared to its previous generation with only one robot arm~\cite{Sharma2019AAS, Beierle2019}, the facility as a whole provides total 13 DOF and allows to take images of the target from the full orientation space and the distance between two objects up to 6 meters. To the authors' knowledge, this capability of TRON is only rivaled by the European Proximity Operations Simulator (EPOS) at DLR, which also  consists of two 6DOF KUKA robot arms to simulate the rendezvous and proximity operations in space\cite{Boge2009EPOS}. In addition, the TRON facility is equipped with 10 Earth albedo light boxes around the room simulating the diffuse light\cite{Beierle2019} and a metal halide arc lamp simulating the direct sunlight. 

The second contribution of this paper is the multi-source calibration procedure of TRON using two independent measurement systems: 1) KUKA telemetry which provides the end-effector poses, and Vicon motion track system which provides the objects' poses via tracking the infrared (IR) markers attached to them. Given rigid fixtures of both the camera and the target onto the respective end-effectors, the calibration amounts to solving for each measurement source the Robot/World Hand/Eye (RWHE) calibration problem\cite{Tabb2017RWHE} associated with either measurement source. This allows one to reconstruct the target's pose based on either KUKA or Vicon measurements. The final pose estimate is obtained by fusing the reconstructed poses in a Bayesian framework, which helps reduce the effect of any bias or noise present in either measurements. Moreover, a criterion is developed to reject any bad measurements reported by Vicon. The result shows that the calibrated poses of the target relative to the camera achieve on average sub-degree orientation and millimeter-level position accuracy at close range.

The third contribution of this paper is the comparative analysis of simulated images under a variety of illumination settings that the TRON facility is capable of recreating using its albedo boxes and the sun lamp. Specifically, a CNN pre-trained on SPEED synthetic training images is used on pairs of synthetic and simulated images with shared pose labels and aligned directions of the light source. A considerable performance gap is observed between two imageries, with far worse performance on the model illuminated with the sun lamp and viewed from certain directions. This suggests that the TRON simulated images exhibit a significant domain gap against the synthetic training images and thus can be used as a good database for validating a CNN's robustness across different domains.

This paper is organized as follows. It first provides an overall description of various components of the TRON testbed. Then, it describes the full calibration procedure and the data fusion mechanism to enable accurate pose label generation. An experiment calibration is run to show the accuracy reported by the testbed, and it ends with a comparative analysis of the synthetic and simulated image qualities using a pre-trained CNN.

\section{Notations}

In this work, $\bm{x}_A \in \mathbb{R}^3$ denotes a 3D vector expressed in a reference frame $A$, and $\bm{x}_A^h = [~\bm{x}_A^\top ~~1~]^\top$ is a homogeneous vector extension of $\bm{x}_A$. Given two reference frames $A$ and $B$, a point $\bm{x}_B$ can be equivalently expressed in $A$ via the following rigid transformation,
\begin{equation} \label{eqn:rigid transformation}
	\bm{x}_A = \bm{R}_{BA}\bm{x}_B + {}^A\bm{t}_{AB},
\end{equation}
where $\bm{R}_{BA} \in \mathbb{R}^{3 \times 3}$ is an orthonormal rotation matrix aligning $B$ to $A$, and ${}^A\bm{t}_{AB}$ is a translation vector from the origin of $A$ to that of $B$ expressed in $A$. Equation \ref{eqn:rigid transformation} is equivalent to the following transformation of homogeneous vectors,
\begin{equation}
	\bm{x}_A^h = \begin{bmatrix} \bm{R}_{BA} & {}^A\bm{t}_{AB} \\ \bm{0}_{1 \times 3} & 1 \end{bmatrix} \bm{x}_B^h = \bm{T}_{BA} \bm{x}_B^h,
\end{equation}
where $\bm{T}_{BA} \in \mathbb{R}^{4 \times 4}$ is a combined roto-translation or transformation matrix between $B$ and $A$.

\section{TRON Facility Description}

\begin{figure}[!t]
	\includegraphics[width=0.9\textwidth]{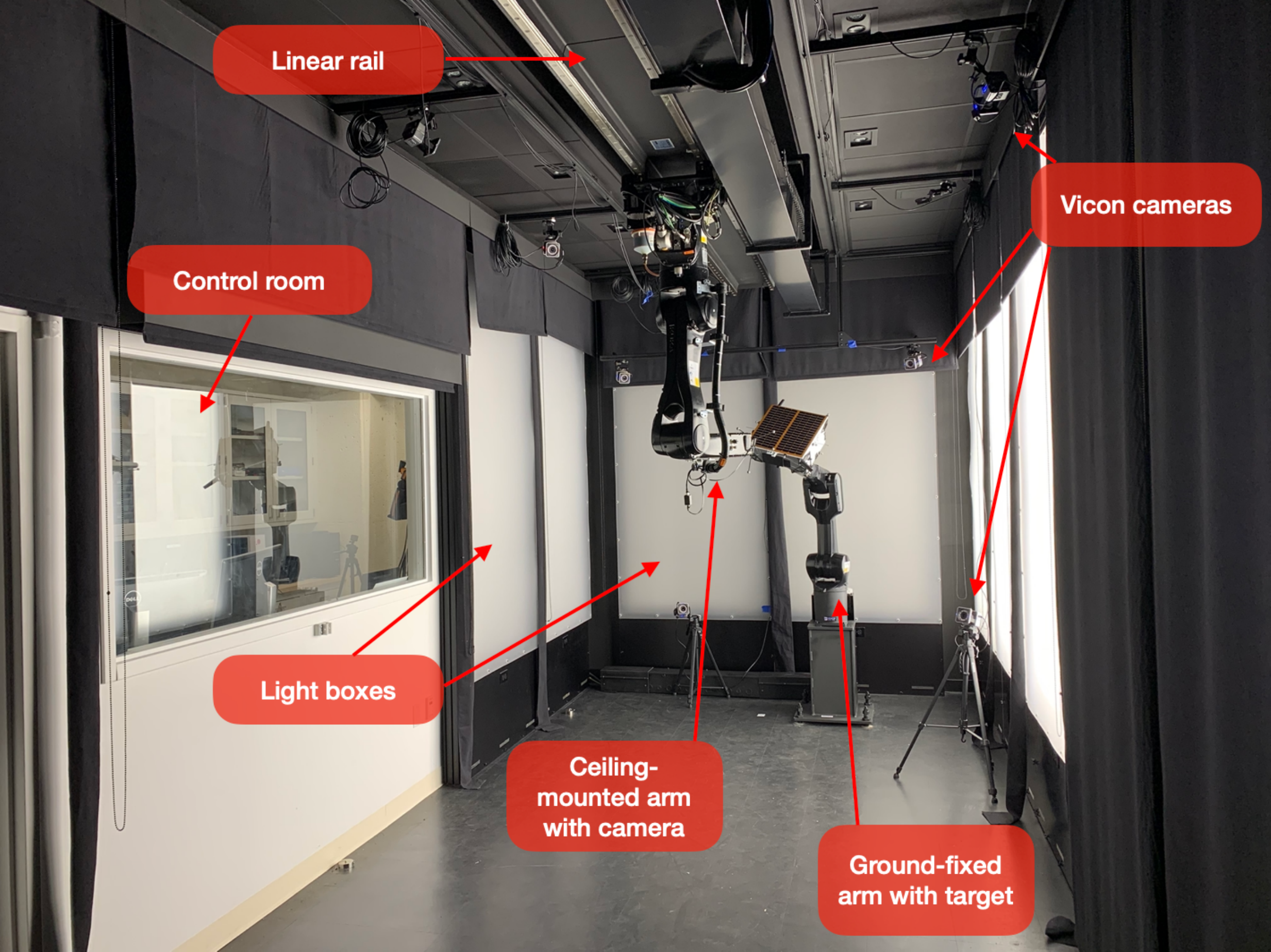}
	\centering
	\caption{TRON simulation room and its components.}
	\label{fig:TRON Simulation Room}
\end{figure}

The TRON facility at the Space Rendezvous Laboratory (SLAB) of Stanford University, visualized in Figure \ref{fig:TRON Simulation Room}, includes a control room and an $8 \times 3 \times 3$ [m] simulation room which consists of various components and machineries to 1) simulate the vision-based rendezvous trajectory of a servicer spacecraft with a camera to its target RSO, and 2) emulate the high-fidelity spaceborne illumination conditions to maximize the realism in the images captured by the camera. This section provides a high-level overview of the components of the facility enabling the above two goals. Note that the facility has seen an extensive upgrade to both its hardware and software capabilities since its original descriptions in Refs [\citen{Sharma2019AAS}], [\citen{Sharma2019PhDThesis}].

\subsection{Pose Reconfiguration \& Annotation}

Given a relative pose to be achieved between the servicer's camera and the RSO model, including satellites, debris, and even celestial bodies such as asteroids and the landing sites, the pose reconfiguration is achieved by simultaneously controlling two 6 DOF KUKA robotic arms\cite{KUKA}, respectively holding a camera and a lightweight, reduced-scale mockup model of the RSO at their end-effectors (see Figure \ref{fig:TRON Simulation Room}). The robot holding a camera is installed onto a ceiling-mounted linear axis rail, providing an additional DOF along the facility and up to approximately 6 meters of separation between the objects along the linear rail. The facility as a whole thus provides total 13 DOF and allows to take images of the target model with the orientation distribution covering the full \textit{SO}(3) space. To the authors' knowledge, this capability is currently unavailable in any other similar testbeds except in the EPOS facility at DLR. The desired RSO's model can be manufactured with two mounting spots at the opposite sides, so that at each mounting configuration, only half of the orientation space is viewed by the camera in order to prevent the robot arm from ever blocking its sight.

In order to track the movements of both objects, the facility includes 12 Vicon Vero cameras that track the IR markers attached to the objects\cite{Vicon}. The Vicon tracker software attaches a reference frame to a set of IR markers associated with each object and is capable of reporting its real-time position and orientation within the facility. Independent from the external measurements provided by Vicon, the KUKA system also provides the telemetry of the poses of both arms' end-effectors in real-time based on their internal joint angles. These two sources of measurements are later jointly used to calibrate the facility, so that a user can retrieve the most accurate estimate of the pose between the camera and the target mockup model in any arbitrary configuration of the robot arms.

\subsection{High-Fidelity Illumination Condition}

\begin{figure}[!t]
	\includegraphics[width=0.8\textwidth]{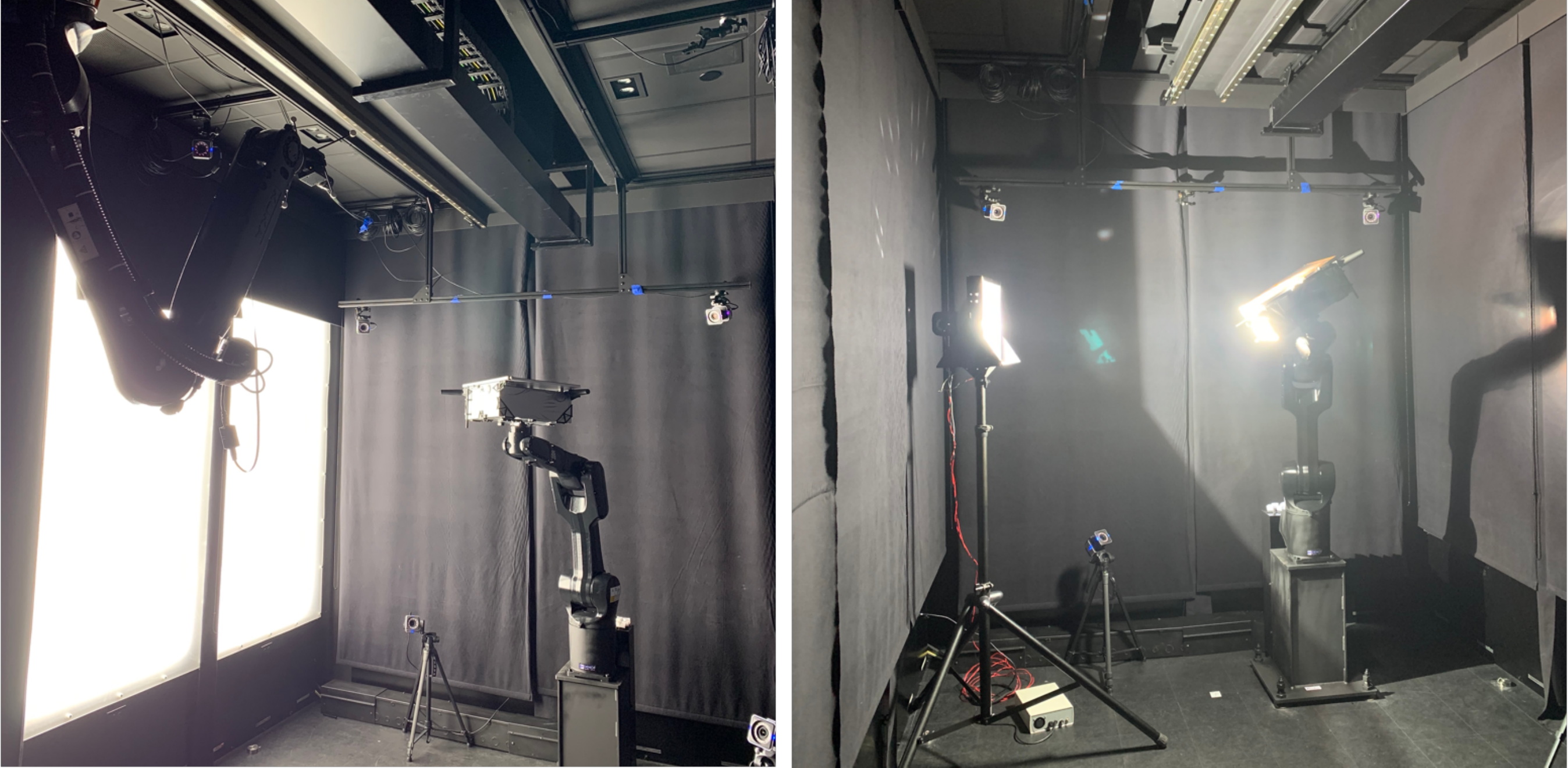}
	\centering
	\caption{(\emph{Left}) Two albedo light boxes activated. (\emph{Right}) Sun lamp activated.}
	\label{fig:TRON Light}
\end{figure}

In order to physically recreate high-fidelity space illumination conditions, the TRON facility is equipped with 10 light boxes around the walls to simulate the diffuse light of Earth albedo\cite{Beierle2019}. A light box consists of a diffuser plate covering hundreds of Light Emitting Diodes (LEDs) arranged in strips that can be regulated in color and intensity. The light boxes are rigorously calibrated to output maximally uniform radiance across the diffuser plates consistent with Earth albedo in Low Earth Orbits (LEO)\cite{LightBox}. The facility also includes a metal halide arc lamp capable of simulating a direct sunlight. Figure \ref{fig:TRON Light} illustrates the operation of both devices and a stark contrast of the effects they cast onto a model. All ambient light sources, including the deactivated light boxes and the windows, are covered with light-absorbing black commando curtains during operations to maximize the effect of diffuse and direct light.

\section{Single-Source Calibration}

The calibration of TRON aims to enable the estimation of the pose between the camera and the target RSO given measurements from KUKA and/or Vicon. This section first formulates the calibration problem based on a set of measurements from a single source: KUKA or Vicon. Then, it provides the descriptions of the full calibration procedure.

\subsection{Reference Frames}

\begin{figure}[!t]
	\includegraphics[width=0.8\textwidth]{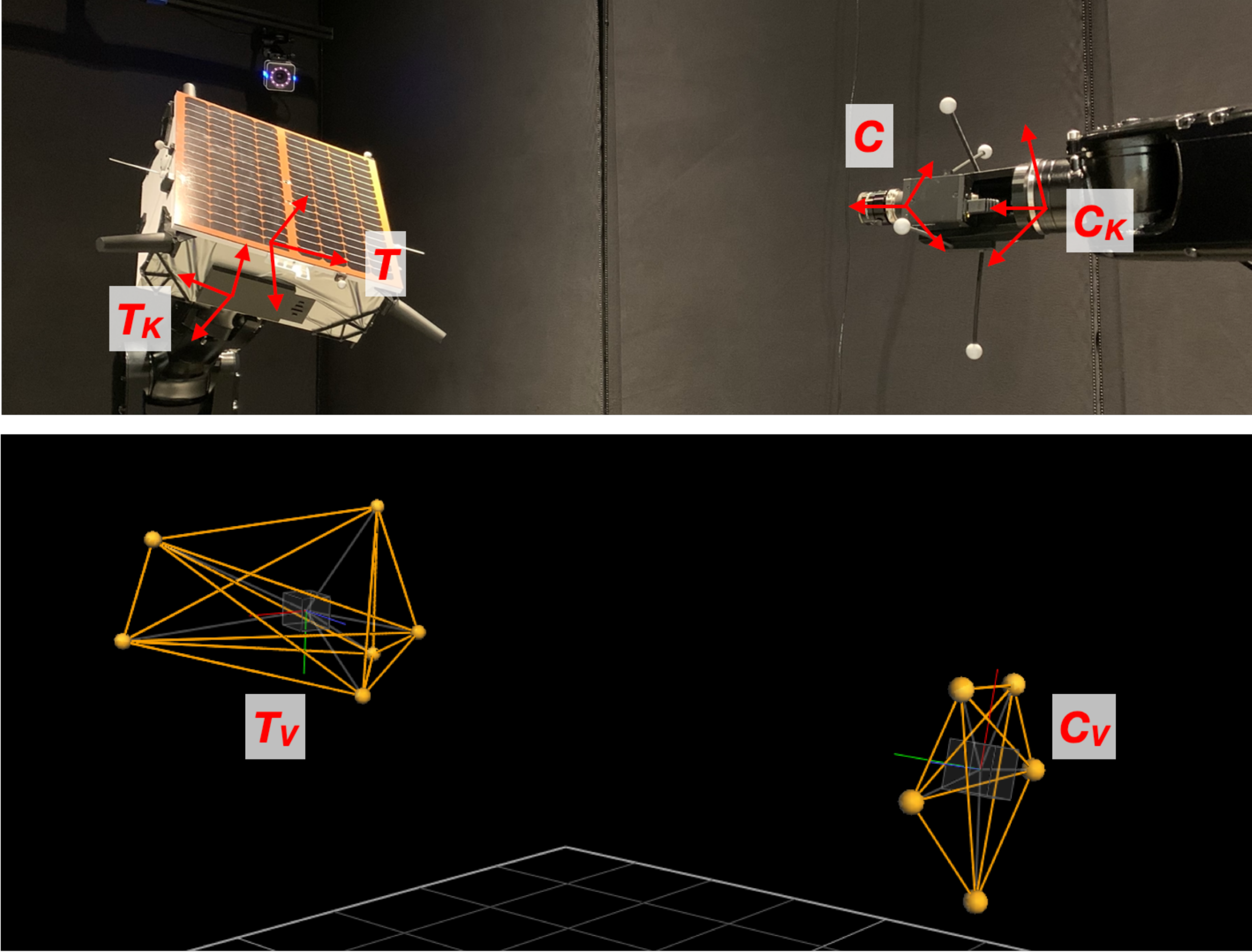}
	\centering
	\caption{(\emph{Top}) End-effector ($C_K, E_K$) and object ($C, T$) reference frames. (\emph{Bottom}) Object reference frames assigned by Vicon ($C_V, T_V$), visualized in the Vicon tracker software as RGB triads.}
	\label{fig:refernce frames}
\end{figure}

The following reference frames used in a single-source calibration are defined below and illustrated in Figure \ref{fig:refernce frames}.

\begin{itemize}
	\item \emph{Camera true reference frame ($C$)}, whose $z$-axis is along the camera boresight and its $xy$-axes form the image plane.
	\item \emph{Target true reference frame ($T$)}, defined according to the 3D model of the target.
	\item \emph{KUKA end-effector frames ($C_K, T_K$)}, defined respectively for the end-effectors of the camera- and the target-holding robot arms.
	\item \emph{Global KUKA reference frame ($K$)}, fixed at an arbitrary location on the ground. 
	\item \emph{Vicon object frames ($C_V, T_V$)}, defined respectively by the Vicon tracker software based on the set of IR markers associated with each object.
	\item \emph{Global Vicon reference frame ($V$)}, also defined by the Vicon tracker software to be fixed at an arbitrary location within the facility.
\end{itemize}

\subsection{Single-Source Calibration Problem}

\begin{figure}[!t]
	\includegraphics[width=0.75\textwidth]{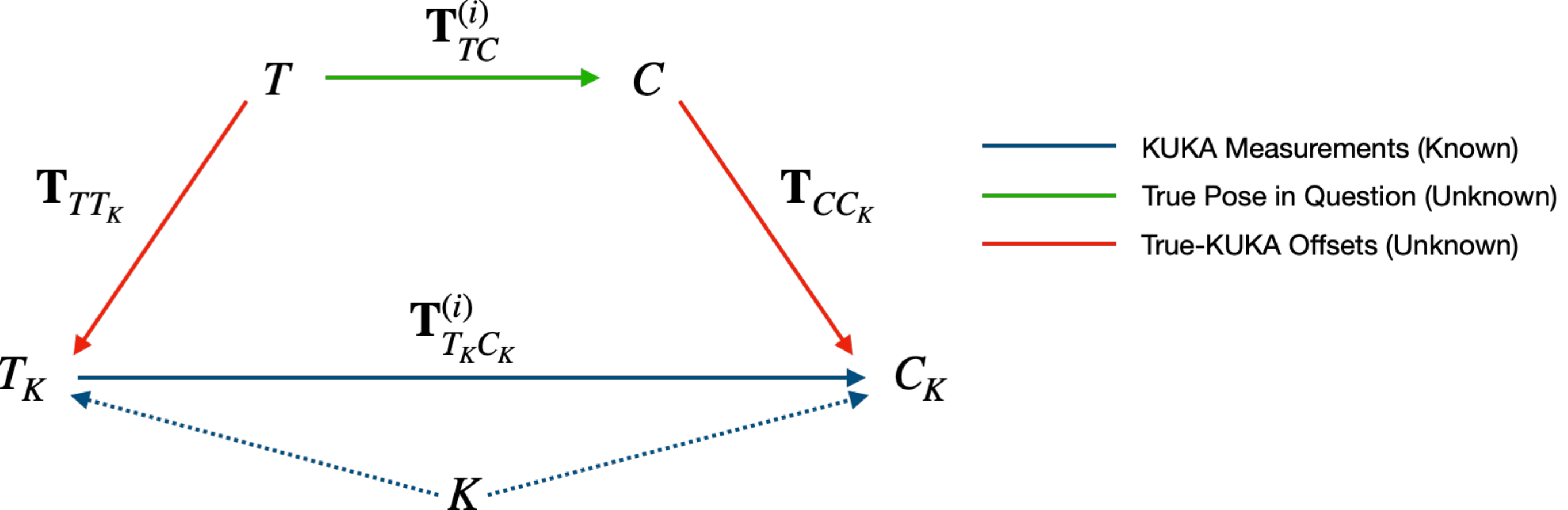}
	\centering
	\caption{Diagram of the KUKA-only calibration problem.}
	\label{fig:kuka_diagram}
\end{figure}

Figure \ref{fig:kuka_diagram} visualizes the calibration problem in a KUKA-only setting. The goal is to estimate the true pose between the camera and the target, $\bm{T}_{TC}$. Note that KUKA provides the measurements $(\bm{T}_{T_K K}, \bm{T}_{C_K K})$, which denote the rigid transformations between the global KUKA frame and both robots' respective end-effector frames. Equivalently, the measurement provides the relative pose between two end-effectors, $\bm{T}_{T_K C_K}^{(i)} = (\bm{T}_{C_K K}^{(i)})^{-1} \bm{T}_{T_K K}^{(i)}, i = 1,\dots,N$, for $N$ calibration data points. There is also an offset between an end-effector and the object it holds, denoted by $(\bm{T}_{T_K T}, \bm{T}_{C_K C})$, that are constant regardless of robot configurations due to rigid fixture of an object onto its end-effector. Then, it is obvious from Figure \ref{fig:kuka_diagram} that if the user has a knowledge of ($\bm{T}_{T_K T}, \bm{T}_{C_K C}$), the true pose in question can be recovered via:
\begin{equation} \label{eqn:single source calibration kuka}
	\tilde{\bm{T}}_{TC, K}^{(i)} = \bm{T}_{C_K C} \bm{T}_{T_K C_K}^{(i)} \bm{T}_{T_K T}^{-1},
\end{equation}
where $\tilde{\bm{T}}_{TC, K}$ indicates that the pose has been estimated based on the KUKA measurements. Note that the setting is completely identical in case of Vicon-only calibration, in which given the measurements $\bm{T}_{T_V C_V}^{(i)}, i = 1,\dots,N$, knowing the constant offsets $(\bm{T}_{T_V T}, \bm{T}_{C_V C})$ allows one to recover the true pose via
\begin{equation} \label{eqn:single source calibration vicon}
	\tilde{\bm{T}}_{TC, V}^{(i)} = \bm{T}_{C_V C} \bm{T}_{T_V C_V}^{(i)} \bm{T}_{T_V T}^{-1},
\end{equation}
where $\tilde{\bm{T}}_{TC, V}$ indicates that the pose has been estimated based on the Vicon measurements. Henceforth, the subscript denoting the measurement source ($K$, $V$) is omitted for notational simplicity unless noted otherwise.
 
The single-source calibration problem now amounts to solving for the constant offsets ($\bm{T}_{T_K T}$, $\bm{T}_{C_K C}$) or ($\bm{T}_{T_V T}$, $\bm{T}_{C_V C}$). Assuming one can recover the true poses for a limited amount of samples in a controlled calibration setting, denoted $\bm{T}_{TC, \textrm{cal}}^{(i)}, i = 1,\dots,N$, then Equations \ref{eqn:single source calibration kuka} and \ref{eqn:single source calibration vicon} can be rearranged to
\begin{equation} \label{eqn:RWHE}
	\bm{T}_{TC,\textrm{cal}}^{(i)} \bm{T}_{T_S T}  = \bm{T}_{C_S C} \bm{T}_{T_S C_S}^{(i)} , i = 1,\dots,N,
\end{equation}
where $S \in \{K, V\}$ denotes the measurement source, and $\bm{T}_{T_S T}, \bm{T}_{C_S C}$ are the only unknowns. Equation \ref{eqn:RWHE} is known as the Robot/World Hand/Eye (RWHE) calibration problem whose solution is well studied in literature\cite{Tabb2017RWHE, Shah2013Kronecker, Li2010RWHEdualquaternion}.

In summary, the calibration procedure to solve the single-source RWHE problem in Equation \ref{eqn:RWHE} consists of two steps: solve for $\bm{T}_{TC,\textrm{cal}}$, then solve the RWHE problem in Equation \ref{eqn:RWHE} for each given measurement source. Next sections describe how these two steps are resolved.

\subsection{Solving $\boldsymbol{T}_{TC,\textrm{cal}}$}

\begin{figure}[!t]
	\includegraphics[width=0.6\textwidth]{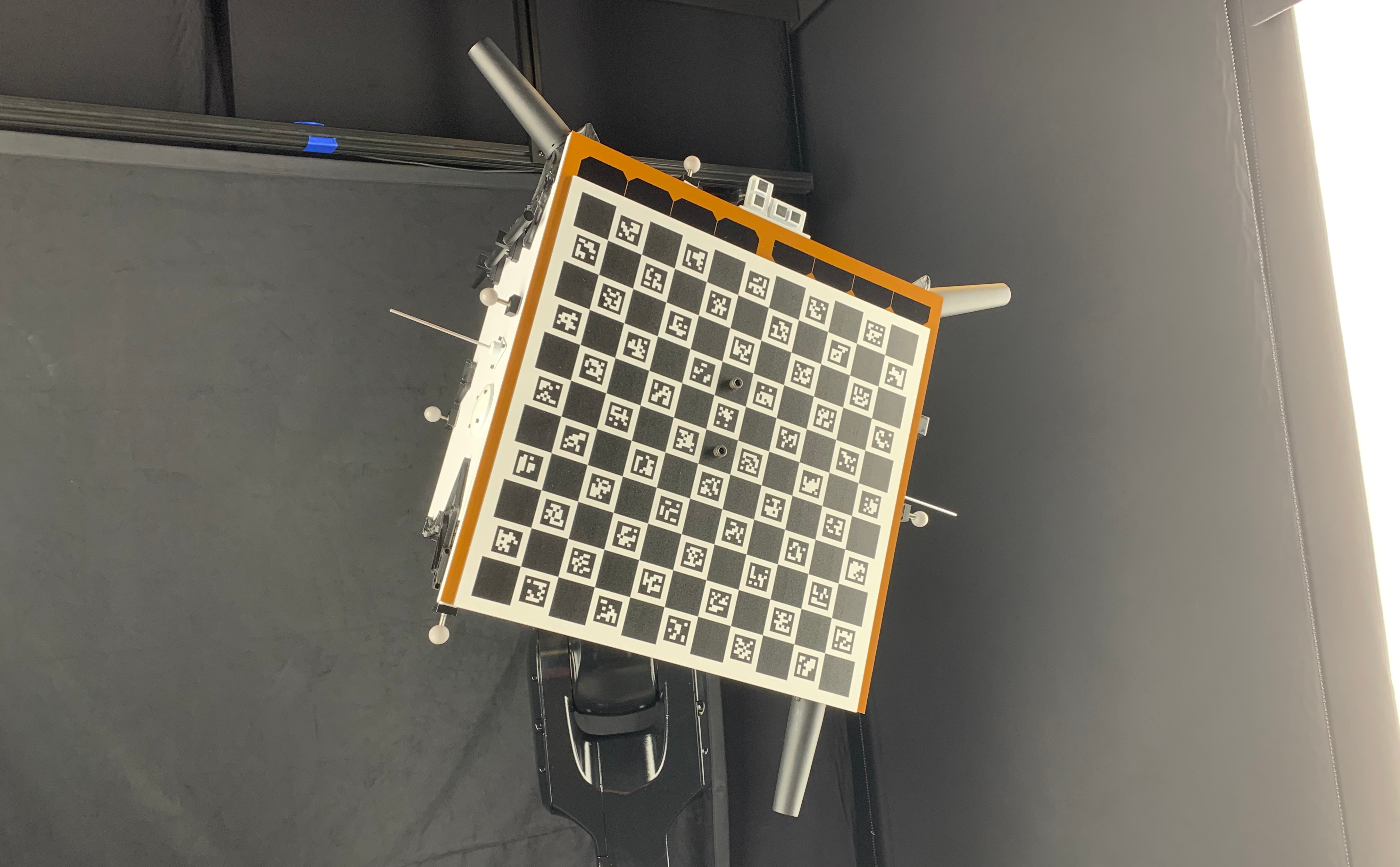}
	\centering
	\caption{Calibration setup with a ChArUco pattern board on the model.}
	\label{fig:charuco setup}
\end{figure}

First, the true pose can be accurately estimated using a known calibration camera and a pattern such as chessboard, asymmetric circle grids, or ChArUco which combines unique ArUco markers\cite{Garrido2014ArUco} and a chessboard. In this work, a ChArUco pattern board is used to solve $\bm{T}_{TC, \textrm{cal}}$. Compared to other options, a ChArUco pattern offers especially flexible choices in terms of its placement with respect to the camera, as the presence of unique ArUco markers allows an easy identification of a partial set of the chessboard corners visible from the images.

As shown in Figure \ref{fig:charuco setup}, a ChArUco pattern board of known dimensions is manufactured and fixed at the known location over a flat surface of the model such as the solar panel, so that the 3D coordinates of the pattern features (i.e., chessboard corners) are known in $T$ reference frame. Then, one can simply solve the Perspective-$n$-Point (P$n$P)\cite{Lepetit2008EPnP} problem to compute $\bm{T}_{TC, \textrm{cal}}^{(i)}$ for $i = 1, \dots, N$, which involves minimizing the following reprojection error:
\begin{equation} \label{eqn:pnp}
	\min_{\bm{T}_{TC}^{(i)}, i=1,\dots,N} \frac{1}{M} \sum_{i = 1}^N \sum_{j = 1}^{M_i} \| \Pi(\bm{T}_{TC}^{(i)}, \bm{X}_j, \bm{K}, \bm{d}) - \bm{x}_j \|^2,
\end{equation}
where $\Pi(\cdot)$ is a projection operator, $\bm{X}_j$ is the 3D location of the $j^\textrm{th}$ feature, $\bm{x}_j$ is the 2D location of the detected $j^\textrm{th}$ feature in the image plane, $\bm{K}$ contains the camera's intrinsic parameters, $\bm{d}$ contains the lens distortion parameters, and $M = \sum_{i=1}^N M_i$ is the total number of visible features in all images. It is also possible to simultaneously perform camera calibration by jointly minimizing ($\bm{T}_{TC}$, $\bm{K}$, $\bm{d}$) in Equation \ref{eqn:pnp}. In this case, the usage of a ChArUco pattern is also advantageous, as the flexibility in its positioning facilitates the detection of the chessboard corners along the edges and corners of the image plane, which increases the quality of the estimated lens distortion parameters.

The detection of the ArUco markers and the chessboard corners is performed using the \texttt{aruco} library of OpenCV 4.5.2\footnote[2]{https://github.com/opencv/opencv\_contrib}. 

\subsection{Solving RWHE Problem}

Once $\bm{T}_{TC,\textrm{cal}}$ are available from solving Equation \ref{eqn:pnp}, the RWHE problem in Equation \ref{eqn:RWHE} for both Vicon and KUKA measurements can be solved using any existing algorithms based on iterative nonlinear least-squares optimization\cite{Tabb2017RWHE}, Kronecker product of matrices\cite{Shah2013Kronecker}, dual quaternions\cite{Li2010RWHEdualquaternion}, and so on. This work frames the RWHE as the following nonlinear least-squares problem\cite{Tabb2017RWHE},
\begin{equation} \label{eqn:rwhe nonlinear}
	\min_{\bm{T}_{T_S T}, \bm{T}_{C_S C}} \sum_{i=1}^N \| \bm{T}_{TC, \textrm{cal}}^{(i)} - \bm{T}_{C_S C} \bm{T}_{T_S C_S}^{(i)} \bm{T}_{T_S T}^{-1} \|_F^2, ~S \in \{ K, V \},
\end{equation}
where $\|\cdot\|_F$ denotes a Frobenius norm. In general, the rotation matrices ($\bm{R}_{T_S T}, \bm{R}_{C_S C}$) can be parametrized using a quaternion, Euler angles, or the matrices themselves can be used directly in the optimization with proper constraints. This work uses the vector representation $\bm{r}$ of a rotation matrix $\bm{R}$ related by the Rodrigues formula, i.e.,
\begin{equation}
	\bm{R}(\bm{r}) = \bm{I}_{3 \times 3}\cos\theta + (1 - \cos\theta)\bm{u}\bm{u}^\top + [\bm{u}]_{\times} \sin\theta,
\end{equation}
where $\theta = \|\bm{r}\|$, $\bm{u} = \bm{r} / \theta$, and $[\bm{u}]_{\times}$ is a skew-symmetric matrix of $\bm{u}$. This parametrization requires no constraint and is also the adopted parametrization in OpenCV, which facilitates the interface with the OpenCV functionalities.

In summary, Equation \ref{eqn:rwhe nonlinear} can be expanded to 
\begin{equation} \label{eqn:rwhe nonlinear expanded}
	\min_{\bm{r}_{TT_S}, {}^{T_S}\bm{t}_{T_S T}, \bm{r}_{C_S C}, {}^C\bm{t}_{C C_S}} \sum_{i=1}^N \| \bm{T}_{TC, \textrm{cal}}^{(i)} - \begin{bmatrix} \bm{R}(\bm{r}_{C_S C}) & {}^C\bm{t}_{C C_S} \\ \bm{0}_{1 \times 3} & 1 \end{bmatrix} \bm{T}_{T_S C_S}^{(i)} \begin{bmatrix} \bm{R}(\bm{r}_{T T_S}) & {}^{T_S}\bm{t}_{T_S T} \\ \bm{0}_{1 \times 3} & 1 \end{bmatrix}  \|_F^2,
\end{equation}
which is solved using the Levenberg-Marquardt algorithm available in MATLAB's \texttt{lsqnonlin} function\footnote[3]{https://www.mathworks.com/help/optim/ug/lsqnonlin.html}. 

\section{Data Fusion}

Once the offsets ($\bm{T}_{T_K T}$, $\bm{T}_{C_K C}$, $\bm{T}_{T_V T}$, $\bm{T}_{C_V C}$) are estimated from the KUKA-based and Vicon-based RWHE calibration problems, one can use either Equation \ref{eqn:single source calibration kuka} or \ref{eqn:single source calibration vicon} to estimate $\tilde{\bm{T}}_{TC}$ for any new data samples. In reality, one can also fuse the results from both measurement sources to help mitigate any bias or noise introduced to the offsets estimated in a single-source measurement setting. In contrast to the single-source calibration, a data fusion approach can take into account the fact that the reference IR markers cannot always be reliably tracked by multiple Vicon cameras, as a number of them inevitably fall into blind spots created by the robot arms and the target model as they move. The condition is exacerbated by the noise introduced by the reflective surfaces of the RSO model and the robot arm components. On the other hand, the KUKA measurements are reported with consistent accuracy based on the internal joint angles of the robot arms regardless of the end-effector pose in the room. Therefore, the limitation of Vicon necessitates a criterion to reject any outlying Vicon measurements, in which case the pose estimated from the KUKA measurement, $\tilde{\bm{T}}_{TC, K}$, is used alone as the final pose label. The following sections describe the data fusion strategy and the rejection criterion.

\subsection{Bayesian Data Fusion}

This work employs a Bayesian approach to data fusion\cite{Kumar2006DataFusion} which utilizes Bayes' theorem and Maximum A Posteriori (MAP) estimation of the posterior state. The Bayesian framework allows one to estimate the true state $X$ as more observations $Z$ become available. Let the probability density function $p(Z~|~X)$ denote the likelihood function encoding the probabilistic information contained in the random variable $Z$ about $X$. By Bayes' theorem, the \emph{a posteriori} probability that $X=x$ given the measurement $Z=z$ can be expressed as
\begin{equation} \label{eqn:Bayes theorem}
	p(X=x~|~Z=z) \propto p(Z=z~|~ X=x) p(X=x)
\end{equation}
where $p(X=x)$ denotes the \emph{a priori} probability that the state $X$ is $x$. If we receive two sets of measurements ($Z_1, Z_2$) that are conditionally independent given the state $X$, i.e., $p(Z=Z_1, z_2~|~X=x) = p(Z=z_1~|~X=x)p(Z=z_2~|~X=x)$, then Equation \ref{eqn:Bayes theorem} expands to 
\begin{equation} \label{eqn:Bayes theorem two meas}
	p(X=x~|~Z=z_1,z_2) \propto p(Z=z_1~|~ X=x) p(Z=z_2~|~ X=x) p(X=x).
\end{equation}
Then, the MAP estimate of the posterior state $\hat{x}$ is given as 
\begin{equation} \label{eqn:maximum a posteriori}
	\hat{x} = \argmax p(Z=z_1~|~ X=x) p(Z=z_2~|~ X=x).
\end{equation}
Assuming the likelihood function is modeled as a Gaussian distribution, i.e., 
\begin{equation}
	p(Z=z~|~X=x) = \frac{1}{\sigma\sqrt{2\pi}} \exp \bigg( \frac{-(x-z)^2}{2\sigma^2} \bigg)
\end{equation}
where the standard deviation $\sigma$ denotes the uncertainty associated with the measurements provided by the sensor, then the MAP estimate of Equation \ref{eqn:maximum a posteriori} becomes
\begin{equation} \label{eqn:bayesian data fusion}
	\hat{x} = \frac{\sigma_2^2}{\sigma_1^2 + \sigma_2^2}z_1 + \frac{\sigma_1^2}{\sigma_1^2 + \sigma_2^2}z_2,
\end{equation}
and the total variance of the fused measurements becomes $\sigma_f^2 = (\sigma_1^{-2} + \sigma_2^{-2})^{-1}$.

Essentially, Equation \ref{eqn:bayesian data fusion} describes the weighted mean of two measurements based on the uncertainties associated with each measurement sources. In the context of TRON calibration, the uncertainty associated with each measurement source is estimated assuming $\bm{T}_{TC, \textrm{cal}}$, the true pose estimated from solving the P$n$P, represents the \emph{mean} transformation. Then, the variance associated with each dimension of the estimated position vector ${}^C\tilde{\bm{t}}_{CT}$ is defined as
\begin{equation} \label{eqn:position variance}
	\sigma_{p}^2 = \frac{1}{N-1} \sum_{i=1}^N \bigg[ \big( {}^C\tilde{\bm{t}}_{CT}^{(i)} \big)_p - \big( {}^C\bm{t}_{CT, \textrm{cal}}^{(i)} \big)_p \bigg]^2,~ p \in \{1, 2, 3\}
\end{equation}
so that they can be applied to Equation \ref{eqn:bayesian data fusion} to compute the dimension-wise weighted mean of two measurements.

To compute the weighted mean in the orientation space, the variance of measurement is taken as a scalar value corresponding to the angular distance between the measured and the mean rotation matrices, i.e., 
\begin{equation} \label{eqn:rotation variance 1}
	d\theta^{(i)} = \arccos \bigg( \frac{\textrm{tr} (d\bm{R}^{(i)}) - 1}{2} \bigg), ~~\textrm{where}~ d\bm{R}^{(i)} = (\bm{R}_{TC}^{(i)})^\top \bm{R}_{TC, \textrm{cal}}^{(i)}.
\end{equation}
Then, the variance associated with the relative orientation is given as
\begin{equation} \label{eqn:rotation variance 2}
	\sigma_{R}^2 = \frac{1}{N-1} \sum_{i=1}^N (d\theta^{(i)})^2.
\end{equation}
Given the uncertainty associated with each measurement, the weighted mean of rotation matrices from KUKA- and Vicon-based RWHE calibrations can be computed by solving
\begin{equation} \label{eqn:data fusion orientation}
	\tilde{\bm{R}}_{TC} = \argmin_{\bm{R} \in \emph{SO}(3)} \sum_{S \in \{K, V\}} w_S\| \bm{R} - \tilde{\bm{R}}_{TC, S} \|_F^2
\end{equation}
where $w_K = \sigma_{R,V}^2 (\sigma_{R,K}^2 + \sigma_{R,V}^2)^{-1}, w_V = \sigma_{R,K}^2 (\sigma_{R,K}^2 + \sigma_{R,V}^2)^{-1}$. The solution to Equation \ref{eqn:data fusion orientation} is given as\cite{Moakher2002Rotation}
\begin{equation}
	\tilde{\bm{R}}_{TC} = \bar{\bm{R}} \bm{U} \bm{D}^{-1/2} \bm{U}^\top
\end{equation}
where $\bar{\bm{R}} = \sum_{S \in \{K, V\}} w_S \tilde{\bm{R}}_{TC, S}$, $\bm{D}$ is the diagonal matrix consisting of the eigenvalues of $\bar{\bm{R}}^\top \bar{\bm{R}}$, and $\bm{U}$ is an orthogonal matrix whose column vectors are the eigenvectors of $\bar{\bm{R}}^\top \bar{\bm{R}}$ corresponding to the diagonal entries of $\bm{D}$. 

\subsection{Rejection Criterion}
As noted previously, it is necessary to develop a criterion based on which one chooses to either reject the Vicon measurement and opt for $\tilde{\bm{T}}_{TC, K}$ or accept the Vicon measurement and perform data fusion. The rejection criterion is established based on the calibration samples which are collected in the optimal setting in terms of the IR marker visibility. Namely, the variance of $\tilde{\bm{T}}_{TC, V}$ with respect to $\tilde{\bm{T}}_{TC, K}$ is computed using an approach similar to Equations \ref{eqn:position variance}, \ref{eqn:rotation variance 1}, \ref{eqn:rotation variance 2} during the calibration. Then, during the future data acquisition step, if the estimated pose $\tilde{\bm{T}}_{TC, V}$ is $1.96\sigma$ away (i.e., 95\% confidence interval) from $\tilde{\bm{T}}_{TC, K}$ in any of the position or orientation components, the Vicon measurement is deemed noisy and thus rejected.

\section{Experiment}

This section first describes the experiment and result of the TRON calibration. Then, a trajectory with 111 pose samples is run to validate that the reported calibration performance extends to an arbitrary relative pose reconfigurable within the simulation room. 

\subsection{Calibration}
The calibration in this work uses the Point Grey Grasshopper 3 camera with a Xenoplan 1.4/17mm lens on a ceiling-mounted robot and a half-scale mockup model of the Tango spacecraft from the PRISMA mission\cite{PRISMA_chapter} on the ground-fixed robot. As visualized in Figure \ref{fig:charuco setup}, the calibration uses a 350 mm $\times$ 350 mm ChArUco board with 11 $\times$ 11 pattern of 30 mm squares printed on a flat aluminum composite panel. The calibration first involves collecting $N = 64$ samples from random orientations up to 45$^\circ$ tilt from a normal vector from the board. The separation between the camera and the board is kept around 0.75 m to ensure consistent accuracy of the detected pattern features. 



The calibration results are reported as an accuracy of the poses estimated from the Vicon- and KUKA-only RWHE calibrations and the data fusion. These estimated poses are compared against those from P$n$P (i.e., $\bm{T}_{TC, \textrm{cal}}$) in terms of three metrics. The mean translation and orientation errors over $N$ samples are reported as
\begin{align} 
	E_\textrm{T} &= \frac{1}{N} \sum_{i = 1}^N \| {}^C\tilde{\bm{t}}_{CT}^{(i)} - {}^C\bm{t}_{CT, \textrm{cal}}^{(i)} \|,\label{eqn:eT} \\ 
	E_\textrm{R} &= \frac{1}{N} \sum_{i=1}^N \arccos \bigg( \frac{\textrm{tr}(\bm{\delta R}) - 1}{2} \bigg), ~~ \bm{\delta R} = (\tilde{\bm{R}}_{TC})^\top \bm{R}_{TC, \textrm{cal}}. \label{eqn:eR}
\end{align}
The third metric is the mean of root-mean-square (RMS) reprojection error of the chessboard corners of the ChArUco board, which is reported as
\begin{equation}
	E_\textrm{p} = \frac{1}{N} \sum_{i = 1}^N \sqrt{\frac{1}{M_i} \sum_{j = 1}^{M_i} \| \bm{\Pi}(\tilde{\bm{T}}_{TC}^{(i)}, \bm{X}_j, \bm{K}, \bm{d}) - \bm{x}_j \|^2}.
\end{equation}

\begin{table}[!t]
	\centering
	\caption{Results of TRON calibration based on single-source measurements and the data fusion technique. Mean values are reported along with one standard deviation errors.}
	\begin{tabular}{@{} r *3c @{}}    
		\toprule
		Metrics & KUKA-only RWHE & Vicon-only RWHE & Data Fusion \\ 
		\midrule
		$E_\textrm{T}$ [mm] & 2.429 $\pm$ 0.866 & 1.208 $\pm$ 0.678 & $\bm{0.815} \pm \bm{0.494}$ \\ 
		$E_\textrm{R}$ [${}^\circ$] & 0.637 $\pm$ 0.120 & 0.172 $\pm$ 0.083 & $\bm{0.169} \pm \bm{0.077}$ \\ 
		$E_\textrm{p}$ [pix] & 5.176 $\pm$ 2.397 & 4.175 $\pm$ 2.718 & $\bm{2.758} \pm \bm{1.716}$ \\ 
		\bottomrule \hline 
	\end{tabular}
	\label{table:rwhe results}
\end{table}

The results of TRON calibration are shown in Table \ref{table:rwhe results}. First, the Vicon-only RWHE is a clear winner compared to KUKA-only RWHE in terms of all metrics, exhibiting millimeter-level position accuracy and millidegree-level orientation accuracy. The KUKA-only RWHE results in a much larger orientation error, which suggests that the fixtures between the end-effectors and the objects, especially the target model, are not perfectly rigid. This is most likely causing the target model to tilt with respect to its end-effector axis up to a couple of millidegrees. However, the data fusion shows improvement over both Vicon- and KUKA-only RWHE, reducing both mean translation and reprojection errors by nearly 33\%. Despite the noticeable discrepancy in performances between KUKA- and Vicon-only RWHE, the performance of data fusion does not deteriorate because it performs a weighted average of $\tilde{\bm{T}}_{TC, V}$ and $\tilde{\bm{T}}_{TC, K}$, with more weights assigned to the variable with less uncertainty. Since the Vicon measurements result in a better calibration performance, the estimated pose from the data fusion is largely close to $\tilde{\bm{T}}_{TC, V}$, with a small correction by $\tilde{\bm{T}}_{TC, K}$ leading to improvement in all metrics.

\subsection{Arbitrary Pose Reconfiguration}

\begin{figure}[!p]
	\includegraphics[width=0.9\textwidth]{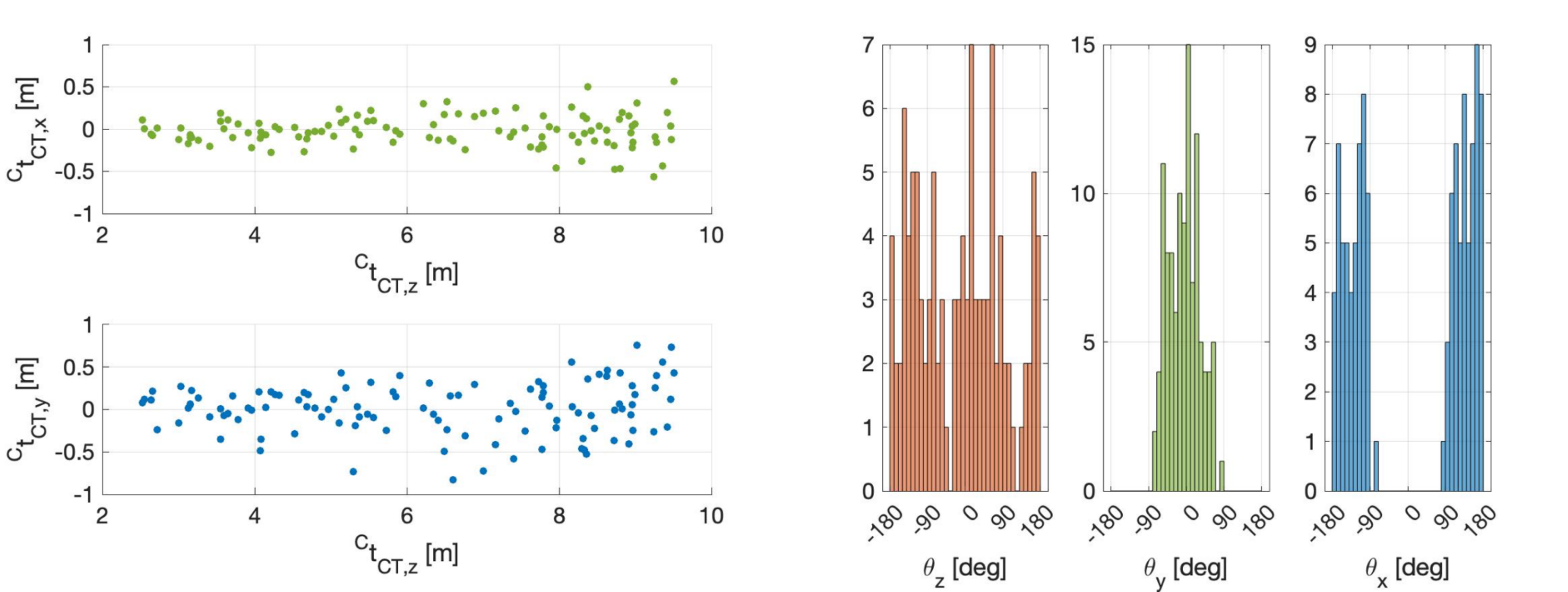}
	\centering
	\caption{Position (\emph{left}) and orientation (\emph{right}) distributions of the test samples. The relative orientation is parametrized as Euler angles for visualization.}
	\label{fig:main_pose_dist}
\end{figure}

\begin{figure}[!p]
	\includegraphics[width=0.85\textwidth]{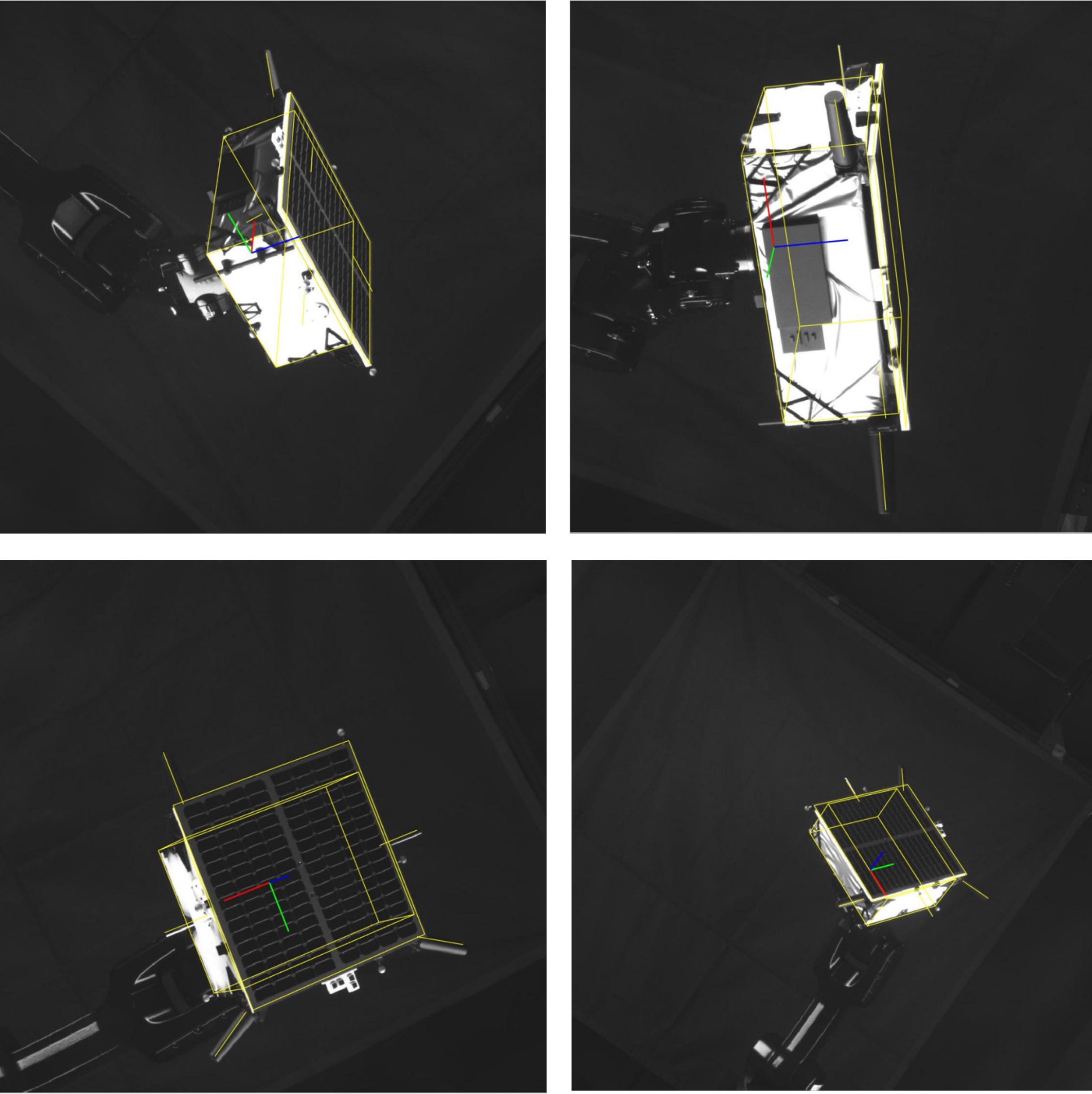}
	\centering
	\caption{Visualization of estimated pose labels from data fusion via projection of the Tango wireframe model.}
	\label{fig:pose_images}
\end{figure}

In order to visually confirm the validity of the calibration and data fusion results for an arbitrary pose reconfiguration, a separate dataset of 111 samples are collected based on the pose distribution visualized in Figure \ref{fig:main_pose_dist}. Note that the relative position assumes a full-scale target, thus allowing a separation along the camera boresight up to nearly 9.5 meters. The select pose labels estimated via data fusion are visualized in Figure \ref{fig:pose_images}, where the Tango spacecraft's full-scale wireframe model is projected onto the images based on the estimated poses. It shows that the calibration and the data fusion method result in consistent accuracy of the estimated pose labels regardless of the camera's orientation and the distance to the target. In general, a similar trend is observed for most of the samples in the dataset to which the data fusion is applicable. It even extends to 15/111 samples in which the Vicon measurements are rejected.

However, it should be noted that there are multiple sources of errors that could spike the translation error up to a centimeter. One is that the model can tilt up to a couple of millidegrees, as evidenced from the calibration result shown in Table \ref{table:rwhe results}. Even such a small error could result in a non-negligible misalignment of the reprojected wireframe model when viewed at a close distance. This would likely happen in case the Vicon measurement, which results in a better orientation estimation, is rejected and must resort to a KUKA measurement. The other is any errors introduced into the orientation offsets present in the camera-holding robot arm, i.e., ($\bm{T}_{C_K C}, \bm{T}_{C_V C}$). In reality, if there is a non-negligible error in either of these orientation offsets that results in deviation of the camera boresight, the resulting error in the target's position would scale with the separation between the camera and the target. For example, if the camera boresight is tilted by 0.1$^\circ$, the target's position would be off by 1.75 cm when placed at 10 meters away. In fact, such an error could be introduced in either KUKA or Vicon measurements as well. While no samples in the collected dataset exhibit such an egregious level of misalignment that could indicate a centimeter-level error, it still remains a possibility.

\section{Comparative Assessment of Image Quality}

This section focuses on the assessment of the qualitiy of images that can be created from the TRON testbed. First, a few select samples with various illumination conditions are showcased to visually demonstrate the range of possible lighting configurations that TRON can simulate using its light boxes and sun lamp. Then, their synthetic counterparts are rendered using the estimated pose labels of the simulated images for a comparative assessment of the domain gap between two imageries. 

\begin{figure}[!p]
	\includegraphics[width=1.0\textwidth]{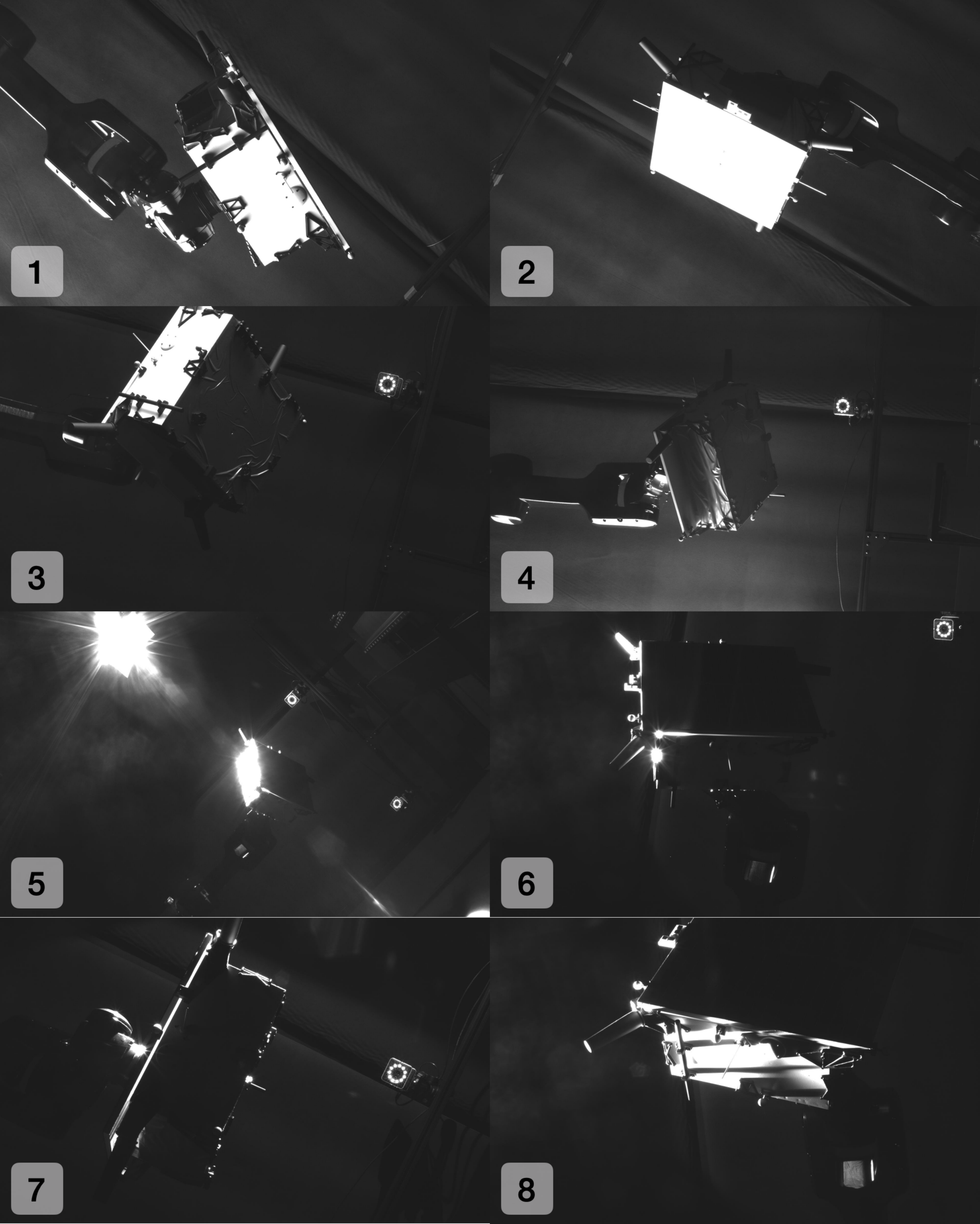}
	\centering
	\caption{Representative illumination configurations that can be recreated in TRON. Images no.~1-4 use the Earth albedo light boxes, and images no.~5-8 use the sun lamp.}
	\label{fig:main images}
\end{figure}

First, Figure \ref{fig:main images} visualizes 8 simulated images with different pose and illumination configurations. The first 4 images (no.~1-4) show the effects of the Earth albedo boxes on the mockup Tango model. As expected from the albedo light, the model is quite evenly illuminated on the parts that face the activated light boxes. One peculiar feature is shown in image no.~2, in which the solar panel completely reflects the light from the albedo boxes it faces due to its high reflectivity. Such effect is not observed in the SPEED synthetic imagery. 

The next 4 images (no.~5-8), on the other hand, show the effects of directed light from the sun lamp cast on the satellite model. Image no.~5 shows some of the boundaries between the model parts disappearing due to extensive directed light and the model surface's reflectivity. It also shows the sun lamp casting a flare effect across the image, as it often happens when the camera is directed near the Sun or any source of direct light. Images no.~6, 7 illustrate the high contrast within the model due to direct sunlight, the feature commonly encountered in spaceborne imageries. Finally, image no.~8 shows the shadow cast by individual parts of the spacecraft, the effect also unobserved in the SPEED synthetic imagery. 

In order to quantitatively evaluate the domain gap between the TRON simulated images of Figure \ref{fig:main images} and its synthetic counterparts rendered from OpenGL, a CNN by Park et al.\cite{Park2019AAS} is pre-trained on SPEED synthetic training set and used as a reference model. Specifically, 495 simulated images are collected based on pose labels sampled from a full orientation space and a position distribution with minimum separation at 6.5 meters in order to align with the distribution of the SPEED synthetic training set\cite{Sharma2019AAS, Sharma2019SPEEDonSDR}. Then, the synthetic images with the same pose labels are rendered with the same setting of the SPEED synthetic imagery and the direction of the light source approximately aligning with the location of the activated albedo boxes and the sun lamp. In order to minimize the distraction from any items in the background, both synthetic and simulated images are masked around the satellite. The CNN performance is measured by the SPEED score defined as
\begin{equation}
	\textrm{SPEED Score} = \frac{1}{N} \sum_{i=1}^N E_\textrm{R}^{(i)} + E_\textrm{T}^{(i)} / \| {}^C\bm{t}_{CT}^{(i)} \|
\end{equation}
where $E_\textrm{R}^{(i)}$ is the rotation error in radians, and $E_\textrm{T}^{(i)}$ is the translation error in meters. Essentially, SPEED score is the average of the sum of the rotation error and the translation error normalized by the norm of the ground-truth translation vector.

\begin{figure}[!t]
	\includegraphics[width=0.7\textwidth]{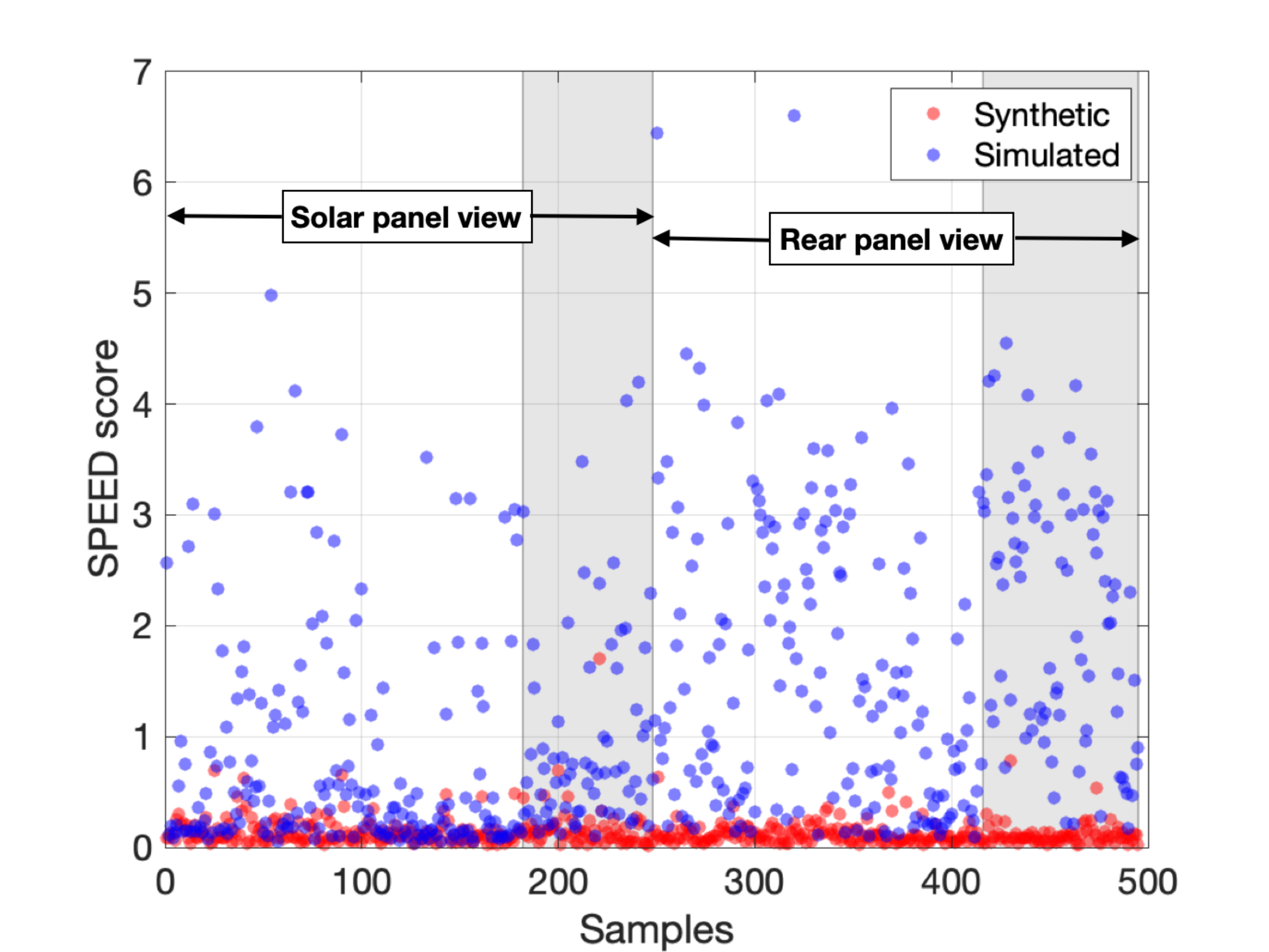}
	\centering
	\caption{SPEED score of the pretrained CNN by Park et al.\cite{Park2019AAS} on synthetic and simulated images of identical pose labels. The first half of the samples have solar panel views, while the other half have rear panel views. The data samples in the shaded region indicate the simulated images are illuminated by the sun lamp. The remaining samples are illuminated by the light boxes.}
	\label{fig:CNN performance}
\end{figure}

\begin{table}[!t]
	\centering
	\caption{Mean SPEED scores for different configurations of the model view and light source.}
	\begin{tabular}{@{} r *2c @{}}    
		\toprule
		Configuration (view, light) & Synthetic & Simulated \\ 
		\midrule
		Solar panel, light boxes & 0.140 & 0.810 \\
		Solar panel, sun lamp & 0.170 & 1.007 \\
		Rear panel, light boxes & 0.121 & 1.568 \\
		Rear panel, sun lamp & 0.114 & 2.062 \\
		\bottomrule \hline 
	\end{tabular}
	\label{table:cnn performance}
\end{table}

\begin{figure}[!t]
	\includegraphics[width=0.7\textwidth]{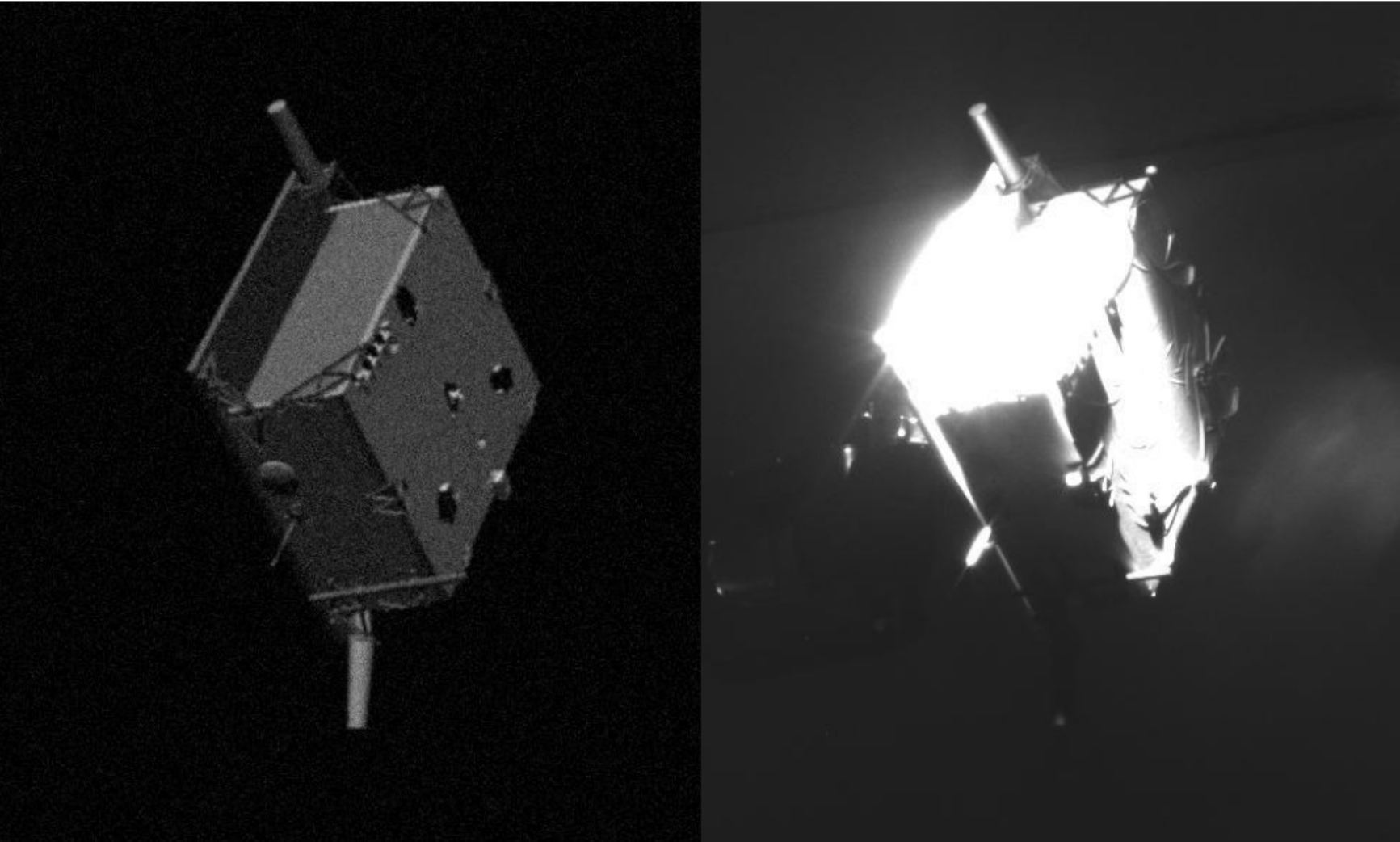}
	\centering
	\caption{Comparison of synthetic vs.~simulated images for rear panel view direction and the sun lamp illumination.}
	\label{fig:image comparison}
\end{figure}

The CNN performance is visualized for all 495 pairs of the synthetic and simulated images in Figure \ref{fig:CNN performance}. It shows in that in general, the SPEED score is worse for simulated images given the same pose labels and the directions of the light source. Table \ref{table:cnn performance} then reports the average SPEED score for different categories of the 495 images. Namely, the collected images have the model mounted in either rear panel or solar panel side, which means the camera has the unobstructed view of the solar panel and rear panel, respectively. The images can also be illuminated by either the light boxes or the sun lamp. The combination of these categories results in 4 different configurations, and the distinctions are made in Figure \ref{fig:CNN performance} as well. Table \ref{table:cnn performance} shows the CNN pre-trained on the SPEED synthetic training set reports on average far worse performance when using the sun lamp and viewing the rear panel of the Tango model. This means not only is the sun lamp casting a more challenging illumination effect, but also the rear panel of the mockup model exhibits a worse gap in the surface texture compared to the synthetic images. Such difference is visualized in Figure \ref{fig:image comparison}, where the synthetic image lacks the texture of the simulated images on its rear panel and retains all of its model features despite the intended illumination of the Sun. This study suggests that the simulated images can be successfully used to measure and validate the domain gap against the synthetic images, and that they are even more powerful when 1) the sun lamp is used for illumination, and 2) the more textured sides of the model are shown. 

\section{Conclusion}

This paper provides a systematic introduction of the next generation Robotic Testbed for Rendezvous and Optical Navigation (TRON) facility at Stanford's Space Rendezvous Laboratory (SLAB), the first-of-a-kind robotic testbed for validating the Convolutional Neural Networks (CNN) or any vision-based navigation algorithms for spaceborne applications. TRON is capable of reconfiguring an arbitrary relative pose between a camera and a mockup model of the target Resident Space Object (RSO) using its two 6 degrees-of-freedom KUKA robot arms, so that one can efficiently generate an arbitrary number of target images with maximally diverse pose distribution. The multi-source calibration of TRON enables the estimation of the true pose between the camera and the target with a millimeter-level position and millidegree-level orientation accuracy. TRON also consists of 10 Earth albedo light boxes and a metal halide arc lamp to emulate diffuse and direct light sources commonly encountered in space missions. The analyses of TRON simulated images with various illumination settings indicate that TRON is capable of generating an imagery that can be used to rigorously validate the robustness of CNNs on an unknown domain of imagery different from the synthetic training images.

Future TRON can benefit from a rigid mounting of a target mockup model and further upgrade to the KUKA robot arms which would drive the position accuracy of each robot to less than half a millimeter. By improving the performance of the KUKA-only calibration, the data fusion can benefit from more equal contributions from both KUKA and Vicon measurements instead of current one-sided dominance from Vicon. This would also remove a rare scenario where a correct pose estimate from a Vicon measurement is rejected in favor of an incorrect pose estimate from a KUKA measurement based on the current rejection criterion.

One of the most important use cases of TRON is the creation of the next generation Spacecraft Pose Estimation Dataset (SPEED), named SPEED+, which will include nearly 10,000 simulated images of the Tango spacecraft with various high-fidelity spaceborne illumination settings and even more synthetic images primarily intended for training. SPEED+ will facilitate the validation of a CNN trained to be robust on an unknown spaceborne domain without access to spaceborne images. Ultimately, SPEED+ will expand to include simulated images of different types of RSOs, such as satellites, space debris, asteroids, etc., so that it can help validate any vision-based navigation algorithms developed for future on-orbit servicing and space situational awareness missions.

\section{Acknowledgement}
The construction of the testbed was partly funded by the U.S. Air Force Office of Scientific Research (AFOSR) through the Defense University Research Instrumentation
Program (DURIP) contract FA9550-18-1-0492, titled High-Fidelity Verification and Validation of Spaceborne Vision-Based Navigation. The authors would like to thank OHB Sweden for the 3D model of the Tango spacecraft used to create the images used in this article.

\bibliographystyle{AAS_publication}   
\bibliography{/Users/taehapark/Documents/reference.bib}    

\end{document}